\documentclass[10pt,twocolumn,letterpaper]{article}

\usepackage[pagenumbers]{cvpr} % To force page numbers, e.g. for an arXiv version

\usepackage{multirow}
\usepackage{booktabs}
\usepackage{subcaption}
\usepackage{setspace}
\usepackage{makecell}
\usepackage{colortbl}
\usepackage[most]{tcolorbox}

\definecolor{cvprblue}{rgb}{0.21,0.49,0.74}
\usepackage[pagebackref,breaklinks,colorlinks,allcolors=cvprblue]{hyperref}
\usepackage[all]{hypcap}

\title{\emph{Sketch-in-Latents}: Eliciting Unified Reasoning in MLLMs}

\author{
	\textbf{Jintao Tong}$^{1}$\quad
	\textbf{Jiaqi Gu}$^{2}$\quad
	\textbf{Yujing Lou}$^2$\quad
	\textbf{Lubin Fan}$^{2\dagger}$\quad
	\textbf{Yixiong Zou}$^{1\dagger}$\\ 
	\textbf{Yue Wu}$^{2}$\quad
	\textbf{Jieping Ye}$^{2}$\quad
	\textbf{Ruixuan Li}$^{1\dagger}$\\
	$^1$School of Computer Science and Technology, Huazhong University of Science and Technology\\
	$^2$Alibaba Cloud Computing~~\quad
	$^{\dagger}$Corresponding authors\\
}

\begin{document}
	
\twocolumn[{%
	\renewcommand\twocolumn[1][]{#1}%
	\maketitle
	\vspace{-3em}
	\begin{center}
		\includegraphics[width=1\textwidth]{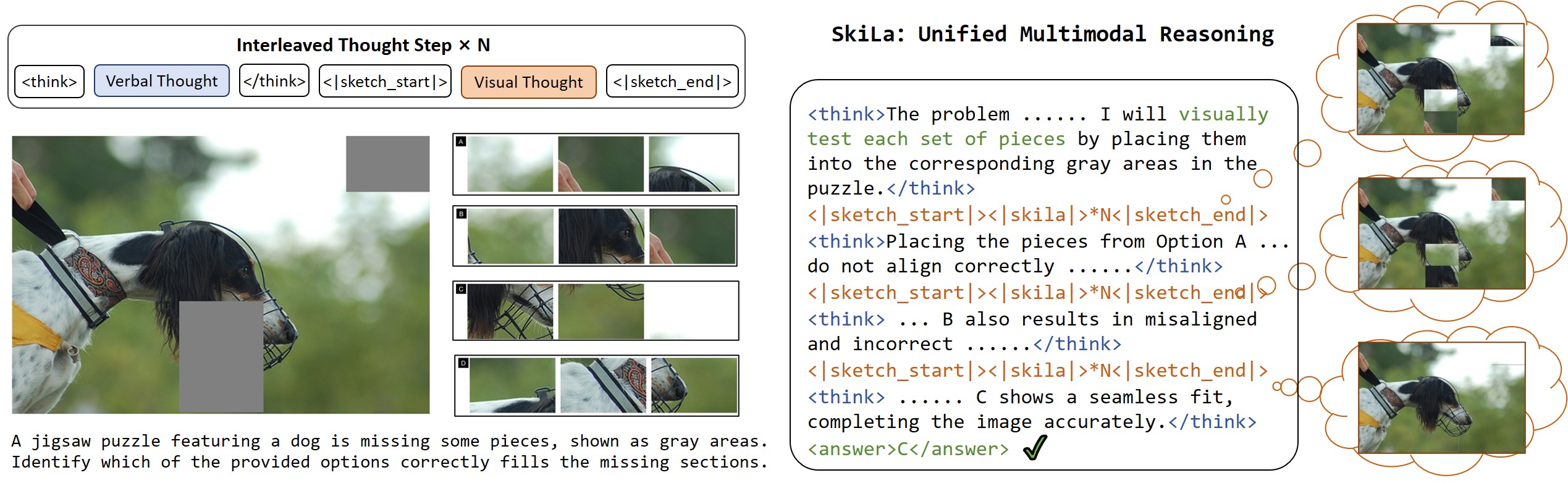}
		\vspace{-1.5em}
		\captionof{figure}{SkiLa empowers MLLMs to think in a unified manner through a hybrid auto-regressive process, generating multi-step, interleaved multi-modal reasoning traces, enabling the model to flexibly think with unpredefined visual-text imagination and interactions. It dynamically alternates between the textual thinking mode to generate textual thoughts and the visual sketching mode to generate latent sketch tokens as visual thoughts, effectively solving challenging tasks that current linguistic-CoT-based methods fail.\vspace{1em}}
		\label{Fig.intro}
	\end{center}
}]

\begin{abstract}
	While Multimodal Large Language Models (MLLMs) excel at visual understanding tasks through text reasoning, they often fall short in scenarios requiring visual imagination. 
	Unlike current works that take predefined external toolkits or generate images during thinking, however, humans can form flexible visual-text imagination and interactions during thinking without predefined toolkits, where one important reason is that humans construct the visual-text thinking process in a unified space inside the brain.
	Inspired by this capability, given that current MLLMs already encode visual and text information in the same feature space, we hold that visual tokens can be seamlessly inserted into the reasoning process carried by text tokens, where ideally, all visual imagination processes can be encoded by the latent features.
	To achieve this goal, we propose Sketch-in-Latents (SkiLa), a novel paradigm for unified multi-modal reasoning that expands the auto-regressive capabilities of MLLMs to natively generate continuous visual embeddings, termed latent sketch tokens, as visual thoughts. During multi-step reasoning, the model dynamically alternates between textual thinking mode for generating textual think tokens and visual sketching mode for generating latent sketch tokens. A latent visual semantics reconstruction mechanism is proposed to ensure these latent sketch tokens are semantically grounded. Extensive experiments demonstrate that SkiLa achieves superior performance on vision-centric tasks while exhibiting strong generalization to diverse general multi-modal benchmarks. Codes will be released at \url{https://github.com/TungChintao/SkiLa}.
\end{abstract}

\section{Introduction}
In recent years, multi-modal large language models (MLLMs)~\cite{bai2025qwen2,liu2023visual,li2023blip,wang2025internvl3} have made remarkable progress by coupling powerful visual encoders with the generative capabilities of large language models (LLMs)~\cite{touvron2023llama,chiang2023vicuna,bai2023qwen}. 
Building on the success of Chain-of-Thought (CoT) reasoning in LLMs~\cite{wei2022chain}, recent multi-modal research~\cite{zhang2023multimodal,huang2025vision,liu2025visual} has adopted this step-by-step reasoning to effectively decompose and solve complex visual-referenced queries.

However, a fundamental limitation persists: the reasoning process of these multi-modal reasoning models is majorly restricted to the linguistic domain~\cite{wang2025vl, yu2025perception}.
Even when processing images, their inference unfolds solely as a sequential generation of text tokens.
This text-centric bottleneck forces them to verbalize problems that humans would solve intuitively through mental visualization.
This limitation stands in sharp contrast to human cognition, where linguistic reasoning is naturally and seamlessly interleaved with internal visual imagination~\cite{farah1985psychophysical,kosslyn1996image}. 
For tasks demanding rich visual-spatial intuition (e.g. jigsaw, object localization)~\cite{fu2024blink}, this limitation becomes particularly pronounced.

Recent efforts to address this gap have explored two main directions. 
The first introduces tool-driven visual exploration~\cite{wu2312v,shao2024visual,zheng2025deepeyes,zhang2025thyme,liu2025visionreasoner} and programmatic visual manipulation~\cite{hu2024visual,zhao2025pyvision}. Tool-based methods enable models to call external utilities for visual interaction, while programmatic approaches generate executable code to perform operations on images, such as drawing auxiliary lines or applying transformations. 
Although an important step, this approach still relies on predefined toolkits or code execution criterion with limited action spaces, unlike humans to flexibly conduct unlimited visual-text imagination or counterfactual visual-text reasoning.
The second direction explores unified foundation models~\cite{team2024chameleon,chen2025janus,deng2025emerging} that generate outputs in both text and images. However, they incur substantial computational costs due to detailed image generation pipelines, which is often constrained by their generative capability. Moreover, their effectiveness has been validated so far only in specific downstream tasks~\cite{li2025imagine,xu2025visual} such as navigation and maze solving, limiting their generalizability.

Unlike current works, one important reason for humans' unlimited visual-text imagination capability is that the visual-text interaction occurs in an internal and unified space inside the human brain.
A fundamental question then emerges: \textit{Can MLLMs transcend their linguistic reasoning framework to generate visual thoughts natively as part of an internal, unified reasoning process}?
This question motivates our goal: enabling MLLMs to internally generate continuous embeddings that natively represent visual thoughts. 
Since in MLLMs, images and texts are all encoded by the LLM into the same feature space, intuitively, to achieve our goal, we hold that \textbf{visual tokens should be able to be seamlessly inserted into the reasoning process carried by text tokens}, forming the internal and unified reasoning process.
Two straightforward advantages of this motivation are: (1) all kinds of visual imaginations can be represented by token features, leading to unlimited visual-thinking manners; (2) the model can seamlessly forming an interleave visual-text reasoning trace, which leads to a more flexible and human-like thinking process.

Based on these insights, drawing inspiration from Coconut~\cite{hao2024training} in the text domain that leverages the LLM's last hidden states as latent text thoughts, we propose Sketch-in-Latents (SkiLa), a novel paradigm for unified multi-modal reasoning that expands the auto-regressive capabilities of MLLMs to natively generate continuous visual embeddings, termed latent sketch tokens, to act as visual thoughts. 
Specifically, SkiLa empowers MLLMs to think in a unified manner through a hybrid auto-regressive process, generating multi-step, interleaved multi-modal reasoning traces. It dynamically alternates between: (1) \textit{Textual thinking mode}: standard next-token prediction on discrete text tokens for textual thoughts, and (2) \textit{Visual sketching mode}: continuous latent sketch token generation for visual thoughts.
To ensure these latent sketch tokens are semantically grounded, we introduce a latent visual semantics reconstruction mechanism. We leverage interleaved text-image reasoning data containing intermediate ``sketch images'' that represent steps in a visual thought process. 
A sketch encoder extracts visual embeddings from these sketches to serve as reconstruction targets. The MLLM is then trained to generate this interleaved sequence: performing standard next-token prediction for textual thoughts and generating latent visual tokens that reconstruct the visual semantics for sketch thoughts.

Extensive experiments validate that SkiLa equips the model with an intrinsic unified reasoning capability. As in Figure~\ref{Fig.intro}, SkiLa can initiate its reasoning linguistically, seamlessly transition to visual sketching mode to perform visual simulations and manipulate visual concepts directly within its latent space, and then revert to text generation. This establishes a novel and flexible paradigm for multi-step unified reasoning where visual cognition and linguistic thoughts are deeply integrated.

\noindent Our contributions are summarized as follows:
\begin{itemize}[label=$\bullet$]
	\item We propose Sketch-in-Latents (SkiLa), a novel unified reasoning paradigm that enables a hybrid auto-regressive process, allowing MLLMs to flexibly and seamlessly interleave multi-step explicit textual thoughts and latent visual thoughts without pre-defined external actions.
	\item We introduce a general-purpose latent visual semantics reconstruction mechanism that trains models to generate meaningful visual representations by reconstructing visual semantics from intermediate sketch images, effectively creating intermediate intrinsic visual thoughts.
	\item We demonstrate our method's effectiveness and generality, showing it not only achieves state-of-the-art results on vision-centric tasks but also exhibits strong generalization and improves performance on diverse general multi-modal benchmarks.
\end{itemize}

\begin{figure*}[!t]
	\centering
	\includegraphics[width=0.96\linewidth]{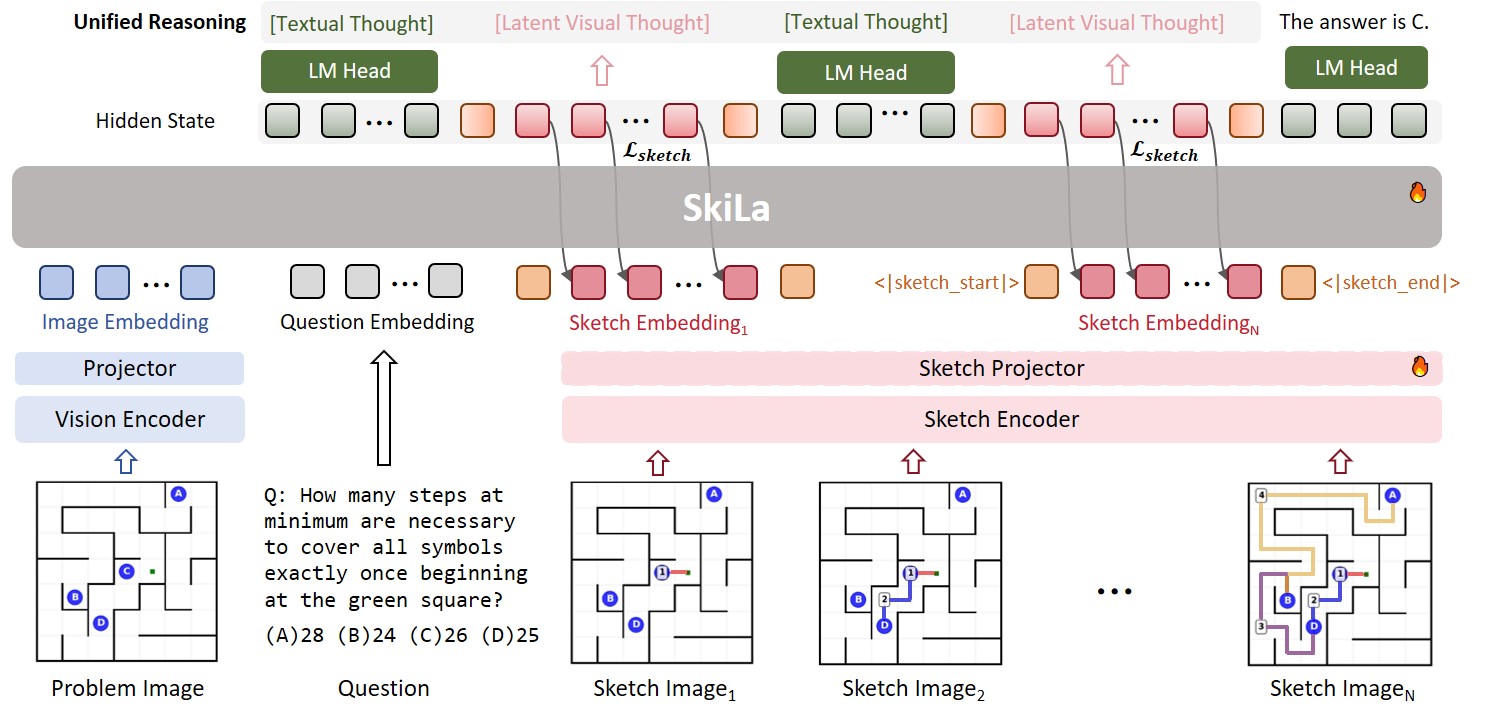}
	\vspace{-0.2cm}
	\caption{The training and inference of SkiLa. It dynamically alternates between textual thinking mode to generate textual thoughts and visual sketching mode to generate latent sketch tokens as visual thoughts. During training, the latent visual semantics reconstruction mechanism leverages a sketch module (encoder and projector), used exclusively during training, to extract visual embeddings from sketch images as reconstruction targets, ensuring the latent sketch tokens are semantically grounded.}
	\label{Fig.overview} \vspace{-0.2cm}
\end{figure*} 

\section{Related Works}

\subsection{Multi-modal Chain-of-Thought}
Chain-of-Thought (CoT)~\cite{wei2022chain} improves LLM's reasoning by decomposing complex problems into intermediate steps. Recent works~\cite{wang2025vl, yu2025perception,yan2025crosslmm, hong2025worldsense, li2025adaptive} have extended this paradigm to multi-modal settings, with methods like Vision-R1~\cite{huang2025vision} and Visual-RFT~\cite{liu2025visual} training MLLMs for step-by-step reasoning. 
However, their reasoning are forced to verbalize visual concepts in the linguistic domain, which hinders their performance on vision-intensive tasks.

Existing efforts to enable ``think with images" fall into two categories. Tool-driven methods~\cite{wu2312v,shao2024visual,zhao2025pyvision} equip MLLMs with predefined external utilities for actions like cropping, drawing, or marking regions. Visual Sketchpad~\cite{hu2024visual}, for instance, generates code to produce auxiliary visual aids, while DeepEyes~\cite{zheng2025deepeyes} generates bounding boxes and then employs zoom-in tools to crop images for better reasoning. 
While effective to an extent, these methods depend on predefined tool-kits with fixed action spaces, limiting flexibility, and failing to cultivate an intrinsic reasoning faculty.
A second direction pursues unified foundation models like Janus-Pro~\cite{chen2025janus} and Bagel~\cite{deng2025emerging}, which interleave text and image generation within reasoning chains.
However, this paradigm is hampered by the prohibitive cost of pixel-level synthesis. Moreover, their understanding is bounded by their generative fidelity, meaning they cannot reason about what they cannot draw. Their effectiveness has also been shown mainly in specific domains such as navigation and maze solving~\cite{xu2025visual,li2025imagine}, raising concerns about their generalizability.
In contrast, SkiLa natively generates intrinsic visual thoughts, eliminating dependencies on external tools or costly pixel-level generation pipelines.

\subsection{Latent Reasoning}
Several studies have explored performing reasoning in a continuous embedding space rather than discrete token space~\cite{hao2024training,cheng2024compressed,shen2025codi,zhang2025soft} to improve the reasoning efficiency of LLMs.
Recent work extends latent reasoning to MLLMs. For instance, Mirage~\cite{yang2025machine} employs hidden states trained to approximate annotated auxiliary visual information, while LVR~\cite{li2025latent} learns to reason over reconstructed image crops within the latent space. However, these methods suffer from two limitations. First, they focus exclusively on visual-only reasoning, failing to interleave it with textual thoughts. 
This separation significantly reduces interpretability because the reasoning process is opaque, providing neither intermediate images for inspection nor corresponding textual explanations.
Second, their reasoning is typically confined to a single inferential step, which limits their applicability to complex, multi-step downstream tasks.
In contrast, our paradigm enables a multi-step, interleaved reasoning trace through a hybrid auto-regressive process, dynamically alternating between textual and visual thoughts.

\section{Methods}
\vspace{-0.1cm}
In this section, we introduce Sketch-in-Latents (SkiLa), a novel reasoning paradigm that enables MLLMs to reason jointly over discrete text and continuous visual representations within a unified auto-regressive process. The overall training pipeline is illustrated in Figure~\ref{Fig.overview}.
Unlike methods that require predefined external tools or image generation pipelines,
SkiLa seamlessly alternates between two distinct operational modes: a textual thinking mode, which generates textual thoughts via standard next-token prediction, and a visual sketching mode, which produces sequences of continuous latent sketch tokens as visual thoughts. 
\subsection{SkiLa: Sketch-in-Latents}
\label{sec:overview}

\subsubsection{Model Architecture}
The SkiLa architecture is built upon a standard MLLM framework, comprising three components: a vision encoder $E_v$ mapping input images $\mathbf{I}$ to visual tokens $\mathbf{V} = E_v(\mathbf{I})$; a modality projector $proj$ aligning $\mathbf{V}$ to the LLM's embedding space ($\mathbf{V}_e = proj(\mathbf{V})$); and an LLM backbone $\theta$ that auto-regressively generates responses conditioned on $\mathbf{V}_e$ and textual embeddings $\mathbf{T}$.

To enable latent sketching, SkiLa introduces a sketch module, which is used exclusively during training. This module, consisting of a sketch encoder $E_s$ and projector $proj_s$, encodes intermediate ``sketch images"(i.e., visual thoughts) $\mathbf{I}_s$ from the training data to generate the visual semantic reconstruction targets $\mathbf{V}_s = proj_s(E_s(\mathbf{I}_s))$.

\subsubsection{Hybrid Auto-Regressive Process}

The SkiLa's LLM backbone is trained to perform a hybrid auto-regressive process to achieve unified internal reasoning. The model generates special tokens, $\texttt{<|sketch\_start|>}$ and $\texttt{<|sketch\_end|>}$, to switch reasoning modes. Upon generating $\texttt{<|sketch\_start|>}$, the model enters ``visual sketching mode'', where it auto-regressively generates a sequence of continuous latent sketch visual tokens $\mathbf{H}_{sketch} = \{\mathbf{h}_1, \dots, \mathbf{h}_K\}$, trained to reconstruct the visual thought semantics $\mathbf{V}_s$. Upon generating $\texttt{<|sketch\_end|>}$ seamlessly resumes standard next-token prediction for generating textual thoughts.

Specifically, given an image-question pair $(\mathbf{I}, \mathbf{Q})$, SkiLa performs reasoning by a hybrid auto-regressive process and gets the final answer. It can generate a unified reasoning trace $\mathcal{S}=\{s_1, \dots, s_N\}$, where 
$$
s_i\sim\mathcal{P}_{\theta}(s|\mathbf{V}_e, \mathbf{T},s_1,\cdots,s_{i-1}), \quad s_i \in \{t_i, h_i\}.
$$ 
$s_i$ can be either text token $t_i$ or latent sketch token $h_i$. Special tokens are omitted from the notation for simplicity.

\subsection{Unified Training Objective}
\label{sec:reconstruction}

Our training objective is to jointly optimize the model's ability to generate coherent text and meaningful latent visual semantics, supervised over an interleaved reasoning trace containing textual thoughts $\mathbf{X}_t$ and sketch images $\mathbf{I}_s$ (i.e., visual thoughts).

For each textual thought $\mathbf{X}_t$, the target is the standard sequence of text tokens $\mathbf{T}_t$. For each sketch image $\mathbf{I}_s$, we formulate a concise visual target. First, the sketch encoder $E_s$ extracts patch-level features, resulting in $M$ tokens (e.g., 729 tokens for SigLIP2). These features are then projected into the LLM's embedding space by the sketch projector $proj_s$. To balance the length of visual thoughts with textual thoughts and to provide concise visual semantics for supervision, we apply average pooling to compress the projected features into a fixed-size sequence of $K$ salient vectors (e.g., $K=27$). The final visual reconstruction target sequence is thus $\mathbf{V}_s = \{\mathbf{v}_{s, 1}, \dots, \mathbf{v}_{s, K}\}$.
We then train the LLM backbone $\theta$ with a two-part objective:

\vspace{0.1cm}
\noindent\textbf{Latent Sketch Reconstruction Loss}\ \ When the model is in latent sketching mode (i.e., \texttt{<|sketch\_start|>} is generated), it is trained to auto-regressively generate a sequence of $K$ latent sketch tokens, $\mathbf{H}_{sketch} = \{\mathbf{h}_1, \dots, \mathbf{h}_K\}$. This is achieved by propagating the last hidden state from the previous step as the input embedding for the current step. We enforce these generated latent sketch tokens to approximate the ground-truth visual semantics $\mathbf{V}_s$ using a mean squared error (MSE) objective:
$$
\mathcal{L}_{Sketch} = \frac{1}{K} \sum_{j=1}^{K} || \mathbf{h}_j - \mathbf{v}_{s, j} ||_2^2
$$
where $\mathbf{h}_j$ is the $j$-th generated latent sketch token and $\mathbf{v}_{s, j}$ is the $j$-th target vector from $\mathbf{V}_s$. The parameters of the sketch projector $proj_s$ are jointly trained via this reconstruction loss, without needing a separate training process.

\vspace{0.1cm}
\noindent\textbf{Next-Token Prediction (NTP) Loss}\ \ For all discrete tokens (including special tokens, textual thoughts $\mathbf{X}_t$, and the final answer $\mathbf{X}_A$), we apply a standard cross-entropy loss. A sketch mask is apply to ensure this loss will be not applied to the latent sketch tokens that enclosed in special tokens \texttt{<|sketch\_start|>}  and \texttt{<|sketch\_end|>}. 
$$
\mathcal{L}_{NTP} = - \sum_{t_i} \log p_{\theta}(t_i | \mathbf{V}_e, \mathbf{T}, s_{1:i}).
$$

\noindent\textbf{Joint Objective}\ \ The final training loss for SkiLa is a weighted sum of these two components, balanced by a hyperparameter $\lambda_{Sketch}$:
$$
\mathcal{L}_{SkiLa} = \mathcal{L}_{NTP} + \lambda_{Sketch} \cdot \mathcal{L}_{Sketch}
$$

\subsection{Inference with Unified Reasoning}
\label{sec:inference}

For inference, the entire sketch module ($E_s$ and $proj_s$) is discarded, incurring zero additional parameter or computational cost compared to a standard MLLM.

Inference proceeds in a hybrid auto-regressive manner as an unified process. The model begins by generating text tokens in response to the input image $\mathbf{I}$ and question $\mathbf{Q}$, sampling from its standard vocabulary. Upon sampling \texttt{<|sketch\_start|>}, the decoding logic switches to ``latent sketching mode". In this mode, the model propagates its last hidden state as the input embedding for the next step, auto-regressively generating a sequence of latent sketch tokens $\mathbf{H}_{sketch}$. This latent generation continues until $\texttt{<|sketch\_end|>}$ is sampled.

During inference, the model may generate more latent sketch tokens than the fixed training length $K$. Therefore, we impose a maximum limit: if the generated sketch tokens exceed twice the training length (i.e., $2K$), the model is forced to exit the latent sketch mode. This prevents excessively long visual thoughts from overshadowing textual reasoning. Upon exiting the latent sketch mode (by sampling $\texttt{<|sketch\_end|>}$ or reaching the limit), the model seamlessly resumes standard next-token prediction. The internally generated latent sketch tokens $\mathbf{H}_{sketch}$ function as the model's visual thought, conditioning subsequent language generation without ever being decoded into pixels.

\renewcommand{\multirowsetup}{\centering}
\definecolor{mygray}{gray}{.92}
\definecolor{ForestGreen}{RGB}{34,139,34}
\newcommand{\fg}[1]{\mathbf{\mathcolor{ForestGreen}{#1}}}
\newcommand{\green}[1]{\textcolor{green!60!black}{#1}}
\definecolor{Forestred}{RGB}{220,50,50}
\definecolor{lightgreen}{rgb}{0.886, 0.941, 0.851}
\newcommand{\fr}[1]{\mathbf{\mathcolor{Forestred}{#1}}}

\begin{table*}
	\centering
	\setstretch{1.1}
	\setlength{\tabcolsep}{4.2pt}
	\resizebox{\linewidth}{!}{
		\begin{tabular}{c |c c c c c c c c |c c c}
			\toprule
			\multirow{2}{*}{\textbf{Methods}} &\multirow{2}{*}{\textbf{MMVP}} &\multirow{2}{*}{\textbf{MMStar}} &\multirow{2}{*}{\textbf{MMBench}} &\multirow{2}{*}{\textbf{RWQA}} &\multirow{2}{*}{\textbf{BLINK}}&\multirow{2}{*}{\textbf{V*}} &\multirow{2}{*}{\textbf{HR$_{4K}$}} &\multirow{2}{*}{\textbf{HR$_{8K}$}} &\multicolumn{3}{c}{\textbf{MME-RealWorld-Lite}} \\
			&&	&&	&  & &	&	&Perception &Reasoning &Overall\\
			\hline
			\rowcolor{mygray}
			\multicolumn{12}{c}{\textit{Close Source Models}} \\
			\hline
			% \Xhline{3\arrayrulewidth} 
			GPT-4o &70.7 &61.6 &82.1 &68.6 &60.0&66.0&59.0&55.5&54.4&48.3&52.0 \\
			GPT-4v &- &56.0&81.0&63.0&58.3&-&-&-&-&-&- \\
			GPT-4o-mini &- &54.8&77.6&67.1&53.6&-&-&-&38.8&35.2&37.4 \\
			Claude3.7-Sonnet &- &65.1&81.4&55.4&56.6&-&-&-&-&-&- \\
			\hline
			\rowcolor{mygray}
			\multicolumn{12}{c}{\textit{Open Source Models}} \\
			\hline
			Gemma3 27B &- &59.6&82.3&62.5&53.5&-&-&-&-&-&- \\
			Cambrian 13B &41.3 &47.1&73.4&58.6&43.1&-&-&-&-&-&-\\
			Janus-Pro 7B &63.3 &46.5&62.6&42.7&38.7&-&-&-&-&-&-\\
			LLaVA-OneVision 7B &-&61.9&83.2&-&53.0&75.4&63.0&59.8&52.8&42.7&48.5\\
			Qwen2.5-VL 7B &66.0 &60.3 &82.5 &67.4 &55.1 &78.5 &68.5 &65.0 &48.6 &37.7&44.3\\
			\midrule
			Direct SFT  &70.2 &61.4&82.1&68.1&56.1&69.1&65.9&61.0&49.6&44.3&47.5 \\
			\midrule
			ROSS 7B~\cite{wang2024reconstructive} &49.3 &53.9&80.5&58.7&45.3&-&-&-&-&-&-\\
			% \rowcolor{mygray}
			% \multicolumn{12}{c}{\textit{Reasoning Methods}} \\
			\cellcolor{green!20}Vision-R1 7B$^*$~\cite{huang2025vision} &72.6 &60.7&79.8&67.1&51.0&80.1&64.8&57.0&49.8&44.9&47.9 \\
			\cellcolor{blue!20}LVR 7B$^*$~\cite{li2025latent} &72.0 &61.3&74.6&67.7&52.5&81.7&69.6&66.1&52.7&43.5&49.1\\
			\midrule
			\cellcolor{blue!20}\textbf{SkiLa-V} &72.0 &64.5&82.4&\textbf{70.5}&56.7&79.6&68.5&65.2&51.7&47.7&50.2\\
			\(\Delta\) (vs Qwen2.5-VL 7B) 
			& \textcolor{red}{+6.0} 
			& \textcolor{red}{+4.2} 
			& -0.1
			& \textcolor{red}{+3.1} 
			& \textcolor{red}{+1.6} 
			& \textcolor{red}{+1.1} 
			& +0.0 
			& \textcolor{red}{+0.2} 
			& \textcolor{red}{+3.1} 
			& \textcolor{red}{+10.0} 
			& \textcolor{red}{+5.9} \\
			
			\midrule
			\cellcolor{red!20}\textbf{SkiLa} &\textbf{75.3} &\textbf{64.8} &\textbf{83.3} &69.3 &\textbf{56.7} &\textbf{84.3} &\textbf{72.0} &\textbf{66.5} &\textbf{56.6} &\textbf{50.2} &\textbf{54.1}\\
			\(\Delta\) (vs Qwen2.5-VL 7B) 
			& \textcolor{red}{+9.3} %\textbf{\green{+9.3}} 
			& \textcolor{red}{+4.5} 
			& \textcolor{red}{+0.8} 
			& \textcolor{red}{+1.9} 
			& \textcolor{red}{+1.6} 
			& \textcolor{red}{+5.8} 
			& \textcolor{red}{+3.5} 
			& \textcolor{red}{+1.5} 
			& \textcolor{red}{+8.0} 
			& \textcolor{red}{+12.5} 
			& \textcolor{red}{+9.8} \\
			\bottomrule
			
		\end{tabular}
		
	}
	\vspace{-0.2cm}
	\caption{\textbf{Experimental results on various vision-centric and multimodal multi-choices tasks.} Our SkiLa with \colorbox{red!20}{``Unified Reasoning''} paradigm outperforms both \colorbox{green!20}{``Textual Reasoning''} and \colorbox{blue!20}{``Visual Reasoning''} paradigms. Here, $*$ denotes the results test by ourselves. }
	\label{table:vision-centric}
\end{table*}

\section{Experiments}
\textbf{Training Details}
We adopt Qwen2.5-VL 7B~\cite{bai2025qwen2} as the base model. The sketch module, which is used only for training-time supervision and discarded at inference, employs a SigLIP2-So/14-384~\cite{tschannen2025siglip} as the default sketch encoder and a single-layer MLP as the sketch projector. The parameters of the base vision encoder, sketch encoder and multi-modal projector are kept frozen, with only the LLM and sketch projector parameters updated. We perform supervised fine-tuning for 1 epoch on 8$\times$A100-80G GPUs with a batch size of 128. A cosine learning rate scheduler is applied with an initial learning rate of 1e-5 for LLM and 1e-4 for sketch projector. 
In addition to \colorbox{red!20}{\textbf{SkiLa}}, trained on interleaved textual and visual thoughts with unified reasoning, we report \colorbox{blue!20}{\textbf{SkiLa-V}}, a purely visual reasoning version trained on the same dataset without textual thoughts.

\vspace{0.2cm}
\noindent \textbf{Training Data}
Our SFT training data is curated from the Zebra-CoT~\cite{li2025zebra}, which is a diverse large-scale dataset with 182K samples, containing logically coherent interleaved text-image reasoning traces. We filter this dataset to select samples suitable for SkiLa training, excluding 3D data and instances where the intermediate sketch images are too complex for the MLLM to effectively learn to reconstruct. Consequently, our training dataset contains 101K samples. The selected data is then reformatted into an interleaved sequence of textual thoughts and visual sketch targets, following the structure shown in Figure~\ref{fig:skila_sample}:
%$\langle\text{think}\rangle$[textual thought 1]$\langle/\text{think}\rangle\langle|\text{sketch\_start}|\rangle$[sketch image 1]$\langle|\text{sketch\_end}|\rangle\langle\text{think}\rangle$[textual thought 2]$\langle/\text{think}\rangle\langle|\text{sketch\_start}|\rangle$[sketch image 2]$\langle|\text{sketch\_end}|\rangle$...$\langle/\text{think}\rangle\langle\text{answer}\rangle$...$\langle/\text{answer}\rangle$.

\begin{figure}[htbp]
	%	\vspace{-0.2cm}
	\footnotesize
	\begin{tcolorbox}[
		colback=gray!5!white,        
		colframe=gray!60!black,     
		coltitle=black,
		colbacktitle=gray!30,       
		title=SkiLa Training Sample,
		fonttitle=\bfseries,
		boxrule=0.8pt,
		enhanced,
		left=2mm, right=2mm, top=1mm, bottom=1mm
		]
		
		\textbf{Interleaved Textual and Visual Thoughts} \\[3pt]
		\textcolor{teal}{\textbf{\texttt{<think>}}} [Textual Thought 1] \textcolor{teal}{\textbf{\texttt{</think>}}} \\[2pt]
		\textcolor{purple}{\textbf{\texttt{<|sketch\_start|>}}} [Sketch Image 1] 
		\textcolor{purple}{\textbf{\texttt{<|sketch\_end|>}}} \\[3pt]
		\textcolor{teal}{\textbf{\texttt{<think>}}} [Textual Thought 2] \textcolor{teal}{\textbf{\texttt{</think>}}} \\[2pt]
		\textcolor{purple}{\textbf{\texttt{<|sketch\_start|>}}} [Sketch Image 2] 
		\textcolor{purple}{\textbf{\texttt{<|sketch\_end|>}}} \\[3pt]
		... ... \\
		\textcolor{teal}{\textbf{\texttt{<think>}}} [Textual Thought N] \textcolor{teal}{\textbf{\texttt{</think>}}} \\[2pt]
		\textcolor{purple}{\textbf{\texttt{<|sketch\_start|>}}} [Sketch Image N] 
		\textcolor{purple}{\textbf{\texttt{<|sketch\_end|>}}} \\[3pt]
		\textcolor{teal}{\textbf{\texttt{<think>}}} [Final Textual Thought] 
		\textcolor{teal}{\textbf{\texttt{</think>}}} \\[2pt]
		\textcolor{black}{\textbf{\texttt{<answer>}}} [Final Answer] 
		\textcolor{black}{\textbf{\texttt{</answer>}}} 
		
	\end{tcolorbox}
	\vspace{-0.3cm}
	\caption{The format structure of the SkiLa training sample.}
	\label{fig:skila_sample}
\end{figure}

\begin{table*}
	\centering
	\setstretch{1.1}
	\resizebox{.9\linewidth}{!}{
		\begin{tabular}{c |c |c c c |c c c c}
			\toprule
			\multirow{2}{*}{\textbf{Methods}} &\multirow{2}{*}{\textbf{TextVQA}} &\multicolumn{3}{c|}{\textbf{MME}} &\multicolumn{4}{c}{\textbf{POPE}}  \\
			&&Perception &Cognition &Overall&Adversarial &Popular &Random &Overall  \\
			\hline
			Qwen2.5-VL 7B &77.5 &1691.4&623.5 &2314.9 &85.5 &86.5 &87.2&86.4\\
			Direct SFT &76.1 &1536.5 &646.8 &2183.3&82.4&83.2&84.4&83.3 \\
			LVR 7B~\cite{li2025latent} &75.6 &1496.1&539.6&2035.7&83.4&84.2&85.1&84.2\\
			\hline
			
			\textbf{SkiLa-V}&80.8&1636.9&646.5&2283.4&85.8&86.7&87.8&86.8\\
			\(\Delta\) (vs Qwen2.5-VL 7B) 
			& \textcolor{red}{+3.3} 
			& -3.2\% %-54.5
			& \textcolor{red}{+3.7\%} %+23
			& -1.3\% %-31.5
			& -0.1 
			& \textcolor{red}{+0.2} 
			& \textcolor{red}{+0.6} 
			& \textcolor{red}{+0.3} 
			\\
			
			\hline
			\textbf{SkiLa} &79.3 &1708.9 &695.4 &2404.3&86.4&87.4 &88.2&87.3\\
			\(\Delta\) (vs Qwen2.5-VL 7B)
			
			& \textcolor{red}{+1.8} 
			& \textcolor{red}{+1.0\%}%{+17.5}
			& \textcolor{red}{+11.5\%}%{+71.9}
			& \textcolor{red}{+3.9\%}%{+89.4}
			& \textcolor{red}{+0.9} 
			& \textcolor{red}{+0.9} 
			& \textcolor{red}{+1.0} 
			& \textcolor{red}{+0.9}
			\\
			\bottomrule
			
		\end{tabular}
		
	}
	\vspace{-0.2cm}
	\caption{Results on VQA and Hallucination Benchmarks. SkiLa demonstrates strong generalization across diverse multi-modal tasks.}\label{table:general-task}
	
\end{table*}

\subsection{Evaluation Benchmarks and Baselines}
\textbf{Evaluation Benchmarks} 
We evaluate SkiLa on a comprehensive suite of vision-centric and multimodal benchmarks to assess its visual perception and reasoning capabilities under diverse settings. For vision-centric evaluation, we adopt MMVP~\cite{tong2024eyes}, RealWorldQA (RWQA)~\cite{grok15v}, BLINK~\cite{fu2024blink}, V*Bench (V*)~\cite{wu2024v}, HRBench (HR$_{4K}$ and HR$_{8K}$)~\cite{wang2025divide}, and MME-RealWorld-Lite~\cite{zhang2024mme}. These benchmarks focus on fine-grained visual perception, visual reasoning, and high-resolution image understanding, covering challenging scenarios such as visual illusions, real-world complexity, and detailed spatial reasoning. For multimodal evaluation, we further test SkiLa on MMStar~\cite{chen2024we} and MMBench~\cite{liu2024mmbench}, which evaluate general-purpose multimodal reasoning in multi-choice settings.

\vspace{0.2cm}
\noindent \textbf{Baselines}
To comprehensively evaluate our approach, we benchmark SkiLa against a carefully selected suite of baselines. First, to isolate the impact of our proposed reasoning framework from the training data itself, we establish a Direct SFT baseline. which is a text-only SFT variant trained on the exact same dataset and with identical parameters as SkiLa. 
We then introduce several state-of-the-art models. For comparison against alternative visual supervision techniques, we include ROSS~\cite{wang2024reconstructive}, which employs a denoising objective to reconstruct the input image, and Janus-Pro~\cite{chen2025janus}, a unified model capable of in-line image generation. For reasoning-centric comparisons, we include Vision-R1~\cite{huang2025vision}, which improves textual CoT with reinforcement learning, and LVR~\cite{li2025latent}, which performs latent-space reasoning by reconstructing the semantics of regions of interest. 
In Table \ref{table:vision-centric}, * denotes the results reproduced by ourselves, while all other results~\cite{achiam2023gpt,liu2024llavanext,team2025gemma} are obtained from the OpenVLM~\cite{duan2024vlmevalkit} and BLINK~\cite{fu2024blink} leaderboards.

\subsection{Main Results}

\subsubsection{Vision-Centric and Multimodal Tasks}
The comprehensive evaluation results on vision-centric and multimodal multi-choice tasks are presented in Table~\ref{table:vision-centric}. Overall, our models consistently outperform the Qwen2.5-VL 7B baseline across all evaluated benchmarks, covering vision-centric perception, high-resolution understanding, and multimodal reasoning tasks. The results highlight some key findings that validate our approach: \textbf{(1) Visual semantics reconstruction boosts overall performance on visually intensive tasks.} A key observation is that for tasks demanding fine-grained visual discernment, our SkiLa-V provides substantially larger gains than other reasoning methods. This suggests that linguistic descriptions are often insufficient to capture the full spectrum of visual information. In contrast, visual semantics reconstruction enables the model to effectively imagine such visual details. For instance, in the BLINK subtask ``Jigsaw Puzzle'', shown in Figure~\ref{Fig.intro}, success hinges on the holistic coherence of the assembled image, which is more natively and effectively represented through visual simulation. \textbf{(2) Unified reasoning sets the best performance by leveraging modality complementarity.} As evidenced by our results, SkiLa consistently achieves best performance, outperforming all single-modality reasoning approaches in our comparison. Notably, it also surpasses leading proprietary models such as GPT-4o and Claude3.7-Sonnet. We attribute this success to our unified paradigm’s ability to harness the complementary strengths of both modalities: it effectively fuses the imaginative, holistic nature of visual thought with the structured, sequential deliberation of textual thought. This outcome highlights the superior scalability and effectiveness of a truly unified reasoning process for MLLMs.

\begin{table*}
	\centering
	\setstretch{1.1}
	\setlength{\tabcolsep}{4.2pt}
	\resizebox{\linewidth}{!}{
		\begin{tabular}{c |c | c c c c c c | c c c}
			\toprule
			\multirow{2}{*}{\textbf{Methods}} &\textbf{Sketch} &\multirow{2}{*}{\textbf{MMVP}} &\multirow{2}{*}{\textbf{MMStar}} &\multirow{2}{*}{\textbf{MMBench}}&\multirow{2}{*}{\textbf{RWQA}} &\multirow{2}{*}{\textbf{BLINK}} &\multirow{2}{*}{\textbf{ V* }} &\multicolumn{3}{|c}{\textbf{MME-RealWorld-Lite}} \\
			&\textbf{Token Num}&&	&&	&  &&Perception	&Reasoning &Overall \\
			\hline
			
			Qwen2.5-VL 7B &- &66.0 &60.3 &82.5 &67.4 &55.1 &78.5 &48.6 &37.7 &44.3\\
			\hline
			\rowcolor{blue!8}
			\hline
			\multicolumn{11}{c}{\textit{Visual Reasoning (SkiLa-V)}} \\
			SkiLa-QwenViT&28 &68.3 &64.4 &81.9 &69.3 &56.2 &79.2 &49.8 &44.9 &47.9\\
			
			SkiLa-CLIP&24&72.0&64.5 &82.5 &70.6&56.0&79.3&50.7&46.1&48.9\\
			
			SkiLa-SigLIP2 &27 &72.0 &64.5 &82.4 &70.5 &56.7 &79.6 &51.7 &47.7 &50.2\\
			\hline
			\rowcolor{red!8}
			\hline
			\multicolumn{11}{c}{\textit{Unified Reasoning (SkiLa)}} \\
			SkiLa-QwenViT&28&73.3 &62.3 &82.8 &68.1 &55.8 &81.5 &52.7&47.3&50.6\\
			
			SkiLa-CLIP&24&73.0&63.4&83.0&68.5&55.5&82.1&52.5&46.7&50.2\\
			
			SkiLa-SigLIP2 &27 &75.3 &64.8 &83.3 &69.3 &56.7 &84.3 &56.6 &50.2 &54.1 \\
			\bottomrule
			
		\end{tabular}
		
	}
	\vspace{-0.2cm}
	\caption{Impact of different sketch encoders for generating latent visual semantics as the SFT reconstruction target.}\label{table:sketch_encoder}
\end{table*}

\begin{figure*}[htbp]
	\centering
	\includegraphics[width=\linewidth]{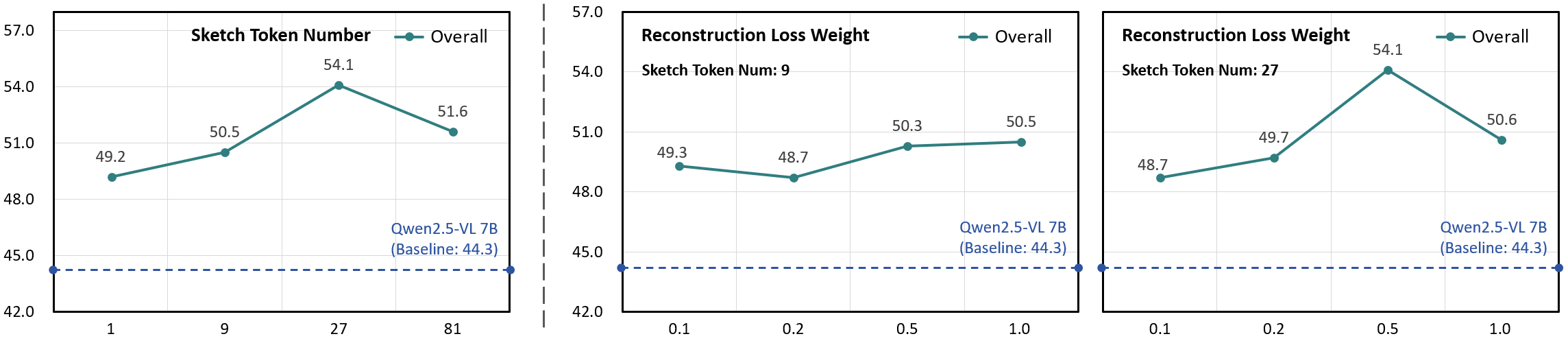}
	\vspace{-0.5cm}
	\caption{Test on MME-RealWorld-Lite. (Left) Impact of reconstructed sketch visual token count on model performance. (Middle and Right) Impact of the latent sketch reconstruction loss weight with 9 and 27 sketch tokens on model performance.}
	\label{Fig.hyper}
	\vspace{-0.2cm}
\end{figure*}

\subsubsection{General Tasks}
Our SkiLa not only achieves state-of-the-art performance on vision-centric benchmarks, but also demonstrates strong generalization across diverse multi-modal tasks. To further assess its generalization, we evaluate SkiLa on several out-of-domain benchmarks, including TextVQA~\cite{singh2019towards}, MME~\cite{fu2025mme}, and POPE~\cite{li2023evaluating}. TextVQA focuses on the integration of textual information within images, testing the model’s ability to align visual and linguistic cues. MME provides a comprehensive assessment of perceptual and cognitive abilities across multi-modal understanding. The POPE benchmark systematically evaluates object hallucination in MLLMs.

As shown in Table~\ref{table:general-task}, while the Direct SFT baseline, trained on the same dataset but without reasoning trajectories, exhibits noticeable performance drops, our method delivers substantial gains. In particular, SkiLa-V and SkiLa demonstrate strong text-grounded reasoning, outperforming the baseline by +3.3 and +1.8 on TextVQA. 
More importantly, SkiLa consistently achieves the best overall performance, enhancing perceptual and cognitive reasoning (+3.9\% on MME) and improving factuality by reducing hallucinations (+0.9 on POPE). These results collectively confirm that SkiLa not only excels in its primary reasoning objectives but also generalizes effectively, enhancing both multi-modal understanding and factual alignment.

\begin{figure*}[!t]
	\centering
	\includegraphics[width=\linewidth]{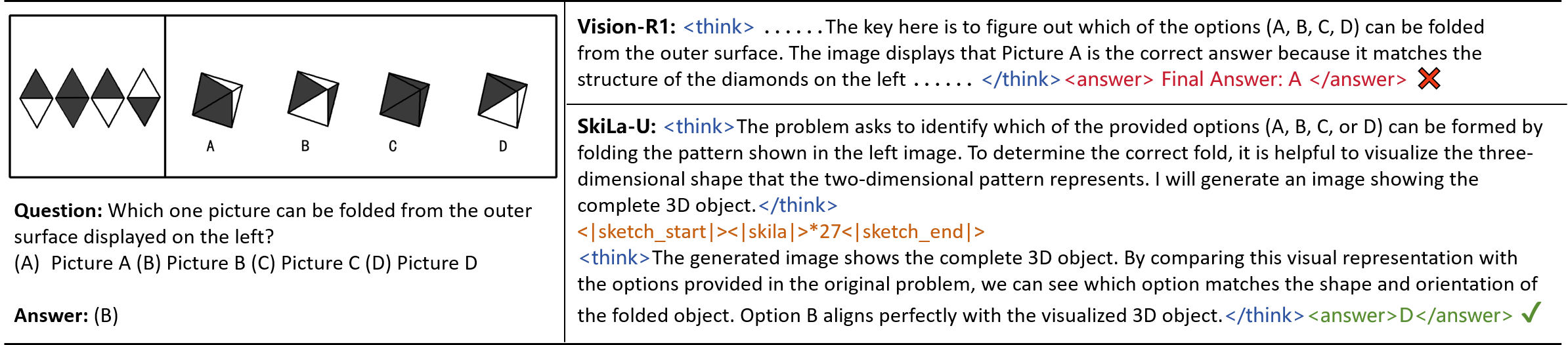}
	\vspace{-0.6cm}
	\caption{An example of spatial imagination. SkiLa generates visual thoughts to imagine the 3D object from a 2D pattern.}
	\label{Fig.case_geo}
\end{figure*}

\begin{figure}[!t]
	\centering
	\includegraphics[width=\linewidth]{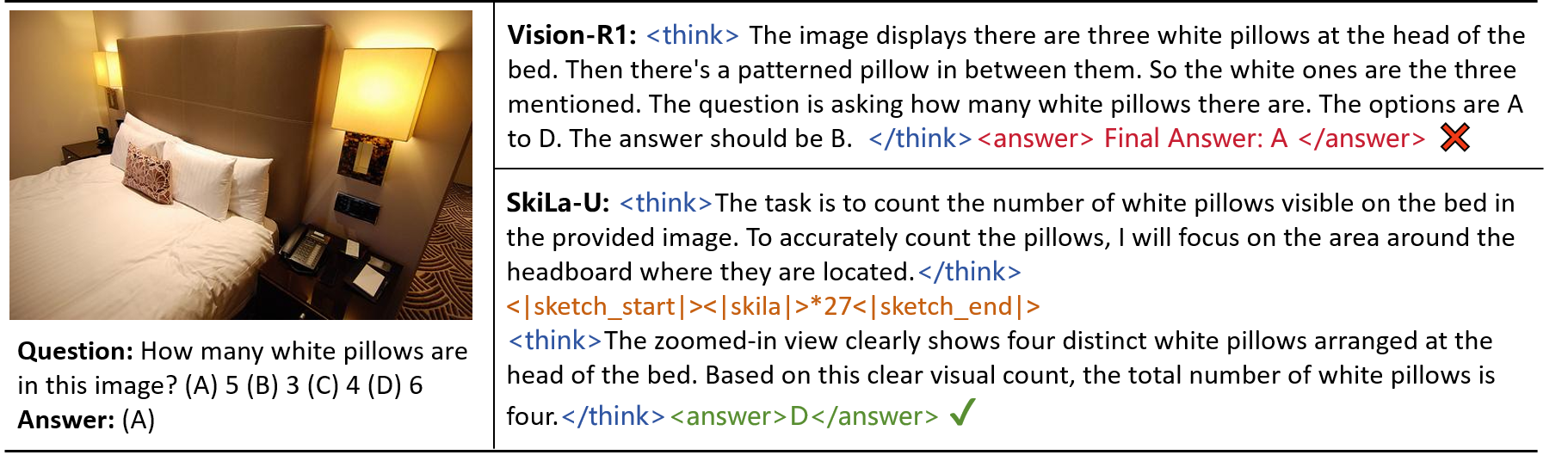}
	\vspace{-0.6cm}
	\caption{An example of hallucination mitigation.}
	\label{Fig.case_count}
\end{figure}

\begin{figure}[!t]
	\centering
	\includegraphics[width=\linewidth]{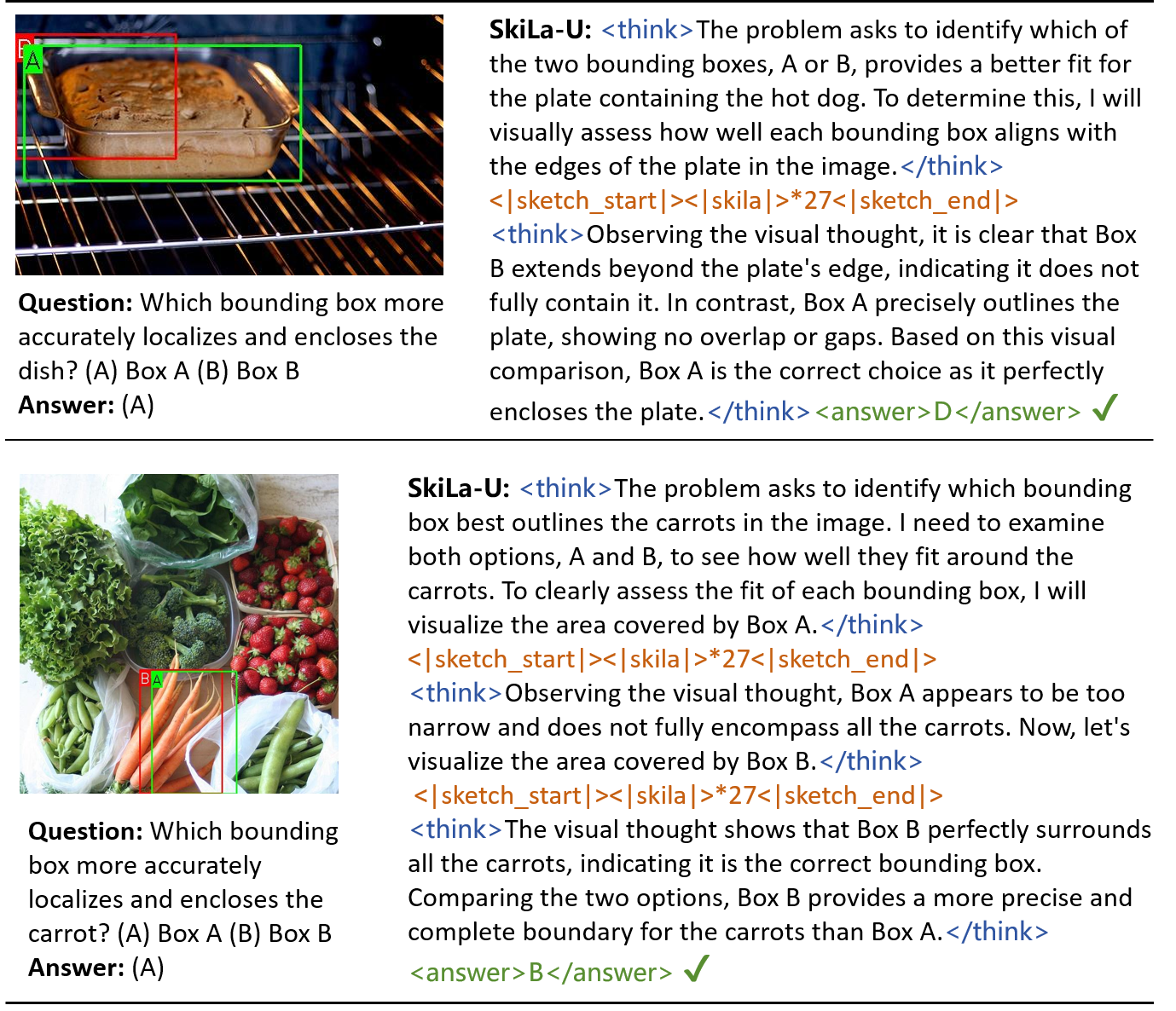}
	\vspace{-0.6cm}
	\caption{An example of adaptive unified reasoning.}
	\label{Fig.case_loc}
\end{figure}

\subsection{Ablation Studies}

\paragraph{Choice of Sketch Encoder}
We investigate the impact of different sketch encoders for generating the visual reconstruction target in Table~\ref{table:sketch_encoder}. The core question is whether the architectural consistency of the native QwenViT is more beneficial than the richer semantic features from powerful, contrastively pre-trained external encoders like CLIP and SigLIP2. To ensure a fair comparison, we select a comparable number of sketch tokens for each: 28 for QwenViT, 24 for CLIP, and 27 for SigLIP2. The results decisively favor external encoders, with SigLIP2 achieving the highest scores in almost benchmarks on SkiLa-V and SkiLa. This demonstrates that the superior feature generality from a powerful, independently trained model is more critical than architectural consistency. Based on its top performance, we adopt SigLIP2 as our default.

\vspace{-0.5cm}
\paragraph{Choice of Sketch Encoder}
We investigate the impact of different sketch encoders for generating the visual reconstruction target in Table~\ref{table:sketch_encoder}. The core question is whether architectural consistency with the native QwenViT is more beneficial than the richer semantic features provided by powerful, contrastively pre-trained external encoders such as CLIP and SigLIP2. To ensure a fair comparison, we select a comparable number of sketch tokens for each encoder: 28 for QwenViT, 24 for CLIP, and 27 for SigLIP2. The results decisively favor external encoders, with SigLIP2 achieving the highest scores on almost all benchmarks for both SkiLa-V and SkiLa. This demonstrates that superior feature generality from a powerful, independently trained model is more critical than architectural consistency. Based on its top performance, we adopt SigLIP2 as our default.

\vspace{0.1cm}
\noindent\textbf{Impact of sketch token number}
We study how the number of reconstructed sketch visual tokens affects model performance. As illustrated in Figure \ref{Fig.hyper} (left), our method consistently outperforms the baseline across a wide range of token counts, from 1 to 81. Performance improves steadily as the number of visual tokens increases, reaching its peak at 27 tokens. Beyond this point, a slight performance drop is observed at 81 tokens. Empirically, 27 tokens yield the best average performance, suggesting that increasing the number of sketch visual tokens does not always lead to further gains. Introducing too many visual tokens may reduce the relative contribution of textual reasoning, highlighting the need to balance visual and textual thoughts. Therefore, to maintain a proper balance between visual and textual thoughts, we choose 27 sketch visual tokens as the default number.

\vspace{0.1cm}
\noindent\textbf{Latent Sketch Reconstruction Loss Weight}
We evaluate the effect of reconstruction loss weights (0.1–1.0) on the model’s average performance across vision-centric tasks, considering two configurations of sketch token numbers, 9 and 27, to analyze their interaction with the loss weight. As shown in Figure \ref{Fig.hyper} (middle), with fewer sketch tokens (9), the model is robust to variations in loss weight, exhibiting only a marginal performance change. With more sketch tokens (27), stronger supervision becomes essential (Figure \ref{Fig.hyper}, right), as insufficient loss weight limits the generation of meaningful sketch tokens and leads to suboptimal performance. These results indicate that the optimal loss weight should align with the number of sketch tokens to ensure effective latent sketch supervision. Based on this, we set 0.5 as the default latent sketch reconstruction loss weight.

\subsection{Discussion}
Beyond quantitative metrics, our evaluation revealed several remarkable emerging capabilities. We now explore these phenomena through detailed case analyses.

\noindent\textbf{Spatial Imagination}\ \ Pure text-based reasoning methods consistently fail on spatial tasks, primarily due to the inherent ambiguity of representing complex 3D geometric properties textually. In contrast, SkiLa overcomes this limitation by generating visual thoughts to mentally simulate the 3D object represented by a 2D pattern, and then uses this mental visualization to compare with candidate options and identify the correct fold, as illustrated in Figure \ref{Fig.case_geo}.

\noindent\textbf{Hallucination Mitigation}\ \ Leveraging our unified reasoning, SkiLa can implicitly zoom in on key image regions to acquire fine-grained information, shown in Figure \ref{Fig.case_count}. This process effectively mitigates hallucinations on visual details, such as correctly identifying all four white pillows, while circumventing the latency associated with complex and explicit tool invocations.

\noindent\textbf{Adaptive Unified Reasoning}\ \ Figure \ref{Fig.case_loc} demonstrates that SkiLa exhibits adaptive unified reasoning capability. The task requires the model to accurately detect and localize the target, and then select the optimal bounding box from two candidates. The upper example shows SkiLa effortlessly identifies the target through a single-step unified reasoning process. In contrast, for a more challenging case, the lower case, where the two candidate boxes exhibit significant overlap and both receive high confidence scores, SkiLa adaptively performs two sequential reasoning steps to discern subtle distinctions and ultimately select the correct box. This illustrates SkiLa’s ability to dynamically adjust its reasoning depth based on task difficulty.

\section{Conclusion}
In this work, we introduced Sketch-in-Latents (SkiLa), a novel unified reasoning paradigm that enables MLLMs to think in a unified manner through a hybrid auto-regressive process. It dynamically alternates between the textual thinking mode to generate textual thoughts and the visual sketching mode to generate latent sketch tokens as visual thoughts. A latent visual semantics reconstruction mechanism is proposed to ensure that the latent sketch tokens are semantically grounded.
Extensive experiments demonstrate that SkiLa achieves superior performance on vision-centric tasks while exhibiting strong generalization to diverse general multi-modal benchmarks. Our future work will develop models that autonomously select the optimal reasoning pattern among textual, visual, and joint, in a task-adaptive manner.
{
	\small
	\bibliographystyle{ieeenat_fullname}
	\bibliography{main}
}

\clearpage
\setcounter{page}{1}
\maketitlesupplementary

\section{Training Data Details}
Our SFT training data is curated from the Zebra-CoT, which is a diverse large-scale dataset with 182K samples, containing logically coherent interleaved text-image reasoning traces. We filter this dataset to select samples suitable for SkiLa training, excluding 3D data and instances where the intermediate sketch images are too complex for the MLLM to effectively learn to reconstruct. Consequently, our training dataset contains 101K samples. Our detailed statistics of training dataset is shown in Table~\ref{tab:sup_dataset}.

\begin{table}[h]
	\centering
	\setstretch{1.1}
	\resizebox{0.85\linewidth}{!}{
		\begin{tabular}{lrr}
			\toprule
			%			\multicolumn{3}{c}{\textbf{SFT training data}} \\
			%			\midrule
			Category & Count & Percentage (\%) \\
			\midrule
			%			\multicolumn{3}{c}{\textit{Visual Reasoning }} \\	
			%			\hline
			Visual Jigsaw & 21,899 &21.6  \\
			Visual Search & 30,000 &29.6  \\
			%			\multicolumn{3}{c}{\textit{Visual Reasoning }} \\	
			%			\hline
			Geometry & 1,058 &1.1  \\
			Checkers & 2,753 &2.7  \\
			Chess & 20,483 &20.2  \\
			Connect Four &2,029 &2.0 \\
			Maze & 20,000 &19.8  \\
			RPM & 3,000 &3.0  \\
			\midrule
			Total &101,222 &100.0 \\
			\bottomrule
		\end{tabular}
	}
	\caption{Detailed statistics of training data.}
	\label{tab:sup_dataset}
\end{table}

\section{Choice of Loss Function}

We evaluate different loss functions for latent sketch reconstruction. Cosine similarity focuses solely on directional alignment between feature vectors, whereas mean squared error (MSE) explicitly minimizes the reconstruction error between generated and target latent sketches. As shown in Table~\ref{tab:sup_loss}, MSE consistently yields higher performance. This is because latent visual representations encode semantics through both direction and magnitude: direction captures coarse alignment in the feature space, while magnitude carries fine-grained visual cues. Cosine similarity, which ignores magnitude differences, may therefore miss subtle but important details. In contrast, MSE preserves both components of the latent features, enabling more precise and faithful visual reconstructions.

\begin{table}[h]
	\centering
	\setstretch{1.1}
	\resizebox{1\linewidth}{!}{
		\begin{tabular}{cccc}
			\toprule
			\multirow{2}{*}{\textbf{Choice of Loss}} &\multicolumn{3}{c}{\textbf{MME-RealWorld-Lite}} \\
			&Perception &Reasoning &Overall\\ 
			\midrule
			cosine similarity loss &56.1 &48.5&53.1\\
			mean squared error loss &56.6 &50.2 &54.1 \\
			\bottomrule
		\end{tabular}
	}
	\caption{Impact of different loss functions for the latent sketch reconstruction loss.}
	\label{tab:sup_loss}
\end{table}

\section{More Cases}
\label{sec:cases}
In the main text, we showed a subset of examples demonstrating SkiLa's effectiveness. Here, we include additional and broader cases across more task types.

\begin{figure*}[!htbp]
\centering
\includegraphics[width=0.96\linewidth]{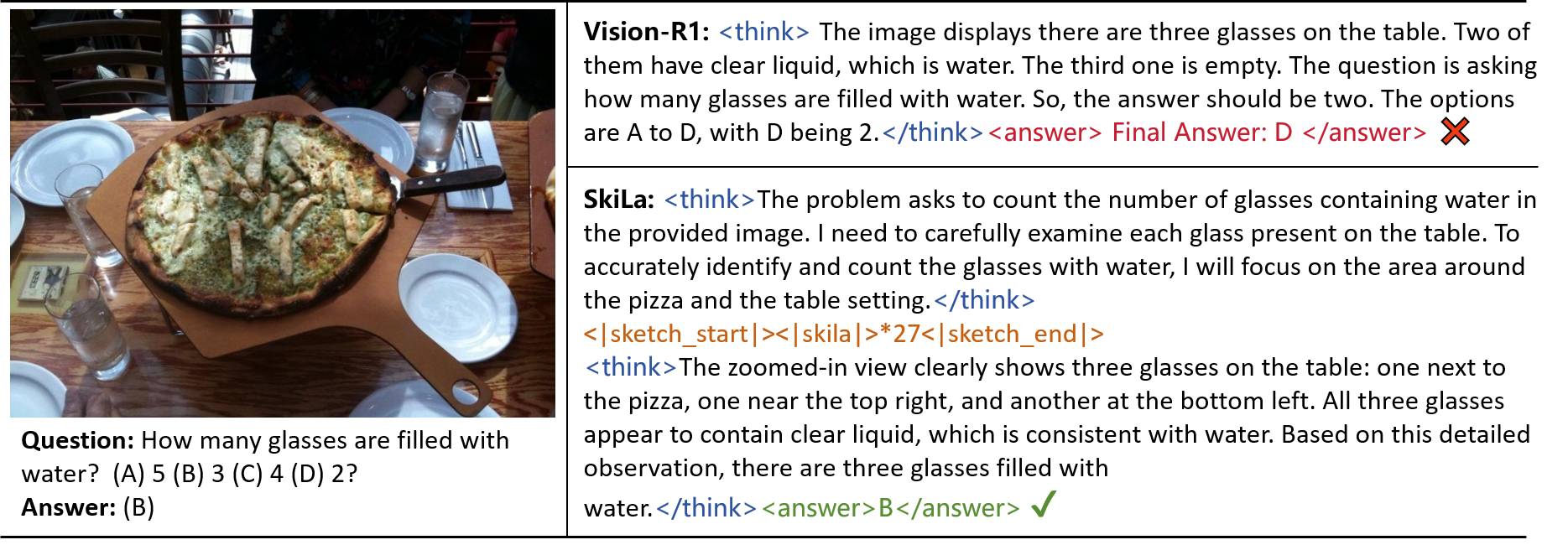}
\caption{A case study demonstrates fine-grained visual grounding capability of SkiLa.}
\label{fig.sup_grounding}
\end{figure*}

\noindent\textbf{Fine-Grained Visual Grounding.} As shown in Figure \ref{fig.sup_grounding}, SkiLa’s unified latent reasoning enables it to internally focus on key regions of an image and recover fine-grained visual details without relying on external tools. This internal refinement allows SkiLa to correctly identify subtle attributes, such as recognizing that all three glasses contain water, while avoiding the latency introduced by explicit region-based operations. Conventional models often overlook such details or produce hallucinated predictions due to insufficient visual grounding.

\begin{figure*}[!htbp]
\centering
\includegraphics[width=0.96\linewidth]{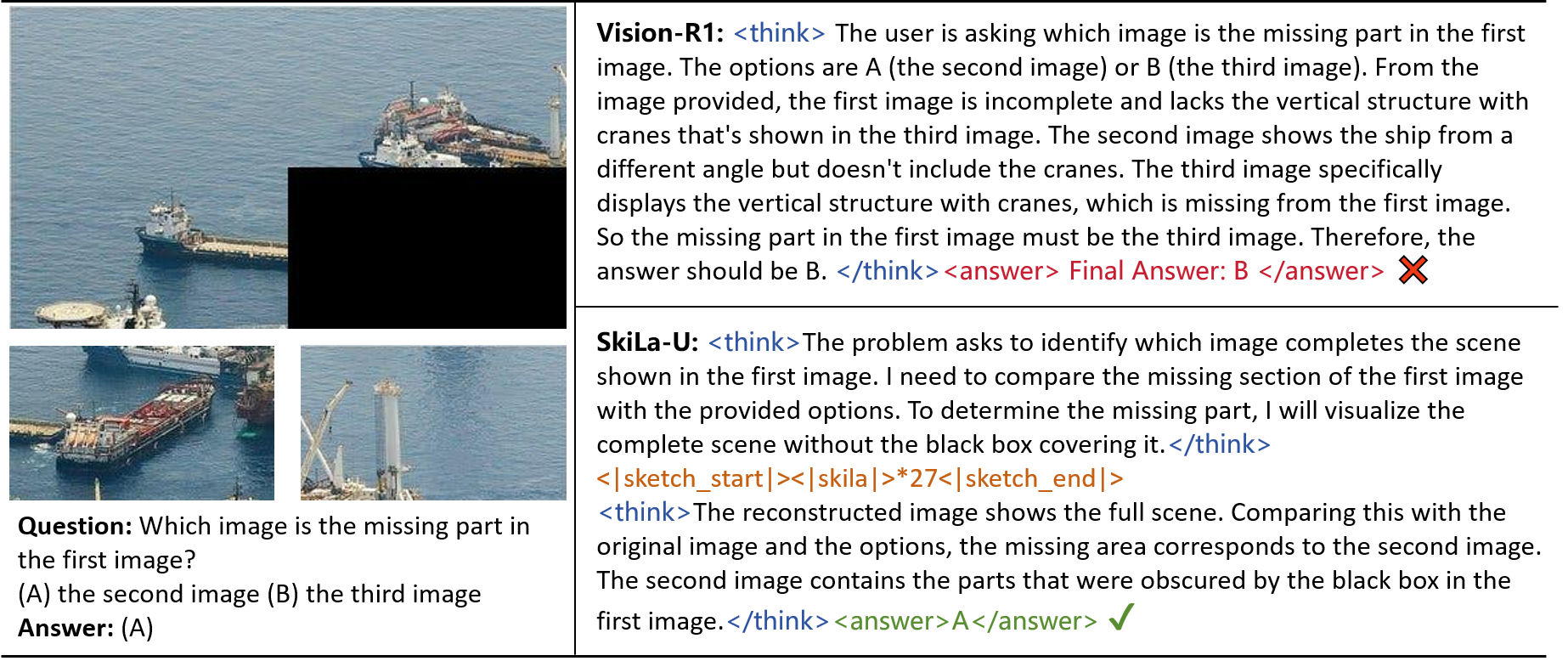}
\caption{A case study demonstrates latent scene completion capability of SkiLa.}
\label{fig.sup_scene}
\end{figure*}

\noindent\textbf{Latent Scene Completion.} As illustrated in Figure \ref{fig.sup_scene}, SkiLa can reconstruct the missing content of a partially occluded image within its latent sketch space and determine which candidate completion is semantically consistent with the underlying scene. This ability to integrate fragmented visual information enables SkiLa to recover the correct jigsaw piece. In contrast, Vision-R1 incorrectly identifies the missing part because it struggles to form a coherent global scene from the incomplete visual input.

\begin{figure*}[!htbp]
\centering
\includegraphics[width=0.96\linewidth]{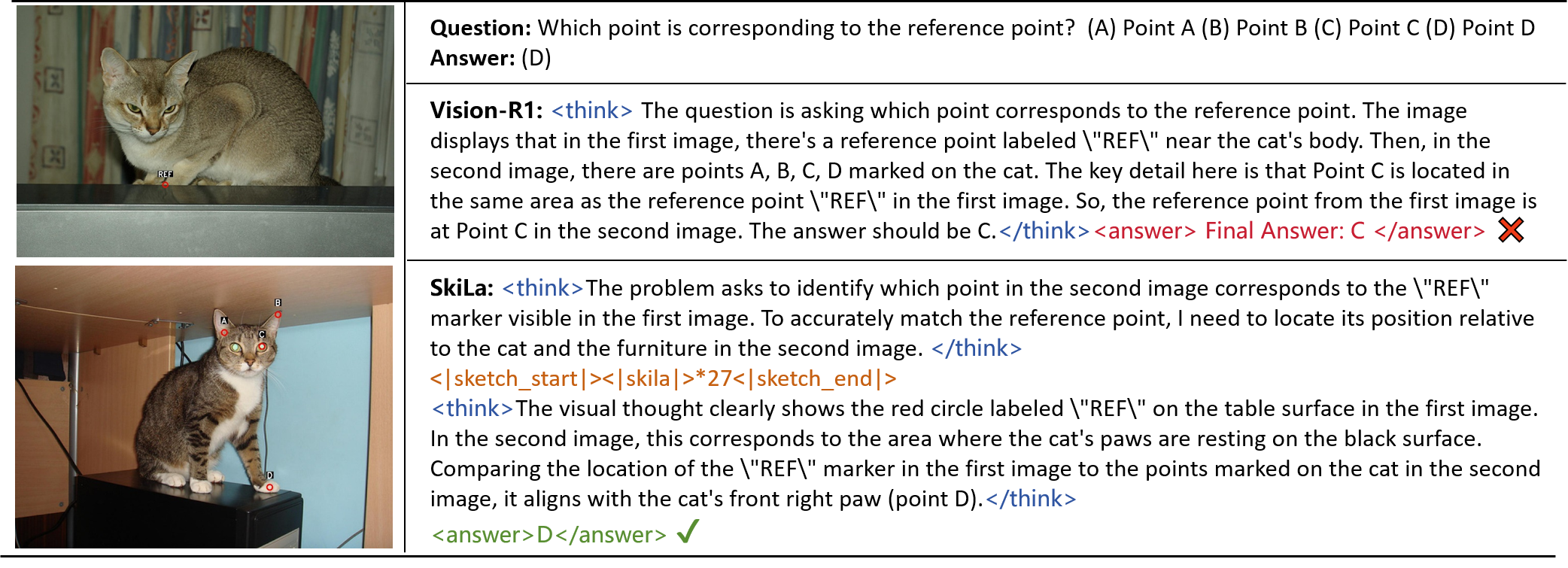}
\caption{A case study demonstrates cross-image visual correspondence alignment capability of SkiLa.}
\label{fig.sup_cross}
\end{figure*}

\noindent\textbf{Cross-Image Visual Alignment.} As shown in Figure~\ref{fig.sup_cross}, SkiLa conducts unified reasoning directly over fine-grained visual semantics. In this example, SkiLa-U accurately aligns the “REF” marker with the cat’s front paw by reconstructing its precise spatial location across images. In contrast, Vision-R1 matches only at a coarse body-level region and therefore predicts incorrectly. This demonstrates that SkiLa not only captures subtle cross-image correspondences but also integrates them seamlessly into its subsequent textual reasoning.

\begin{figure*}[!htbp]
\centering
\includegraphics[width=0.96\linewidth]{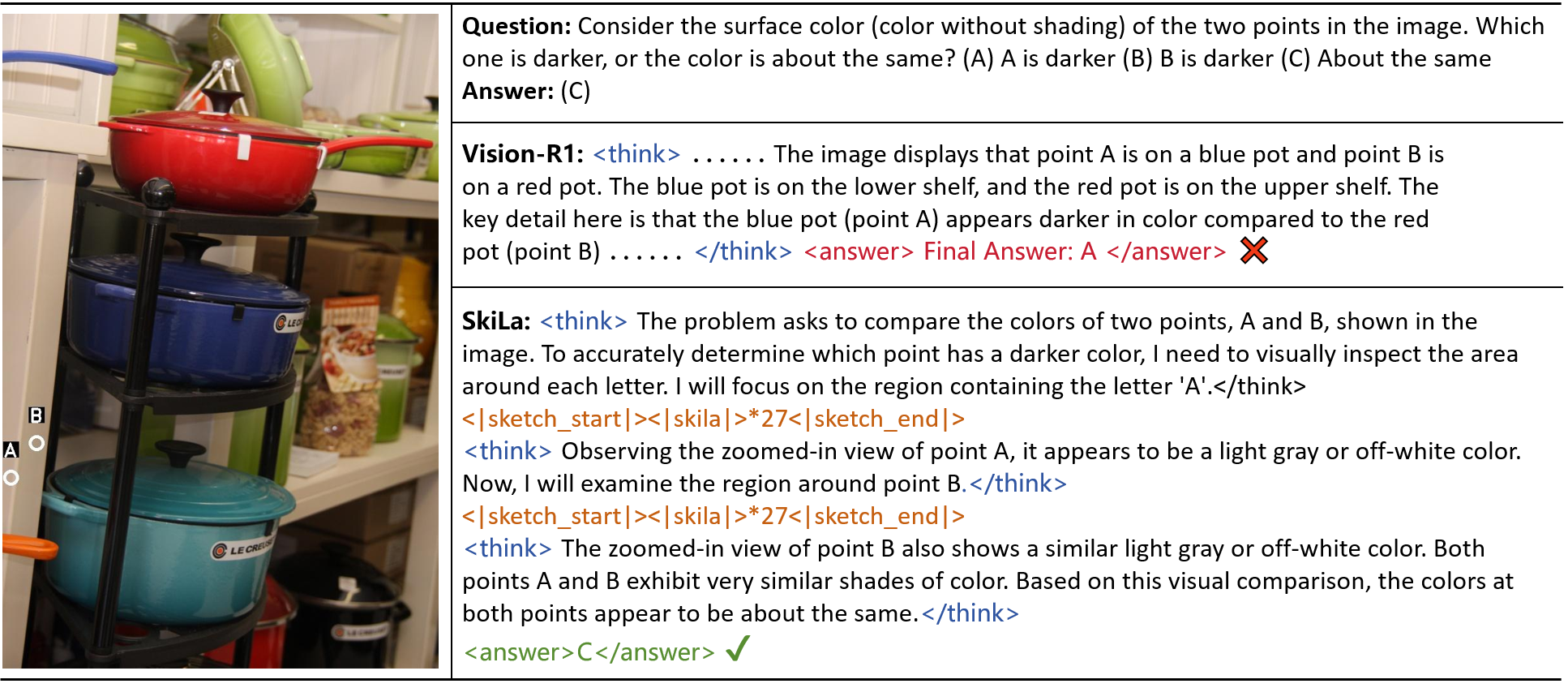}
\caption{A case study demonstrates multi-round visual refinement capability of SkiLa.}
\label{fig.sup_multi}
\end{figure*}

\noindent\textbf{Multi-Round Visual Refinement.} SkiLa demonstrates the ability to perform detailed, multi-step visual reasoning by iteratively examining and comparing regions of interest. As shown in Figure~\ref{fig.sup_multi}, it inspects the two points across multiple rounds, progressively focusing on their local visual properties. By integrating observations step by step, SkiLa accurately determines that the points have similar colors. In contrast, Vision-R1 relies on a coarse global judgment and predicts incorrectly. This case illustrates how SkiLa can conduct stepwise, focused analysis to capture subtle visual features that conventional models often miss.

\end{document}